# Human Expression Recognition using Facial Shape Based Fourier Descriptors Fusion

Ali Raza Shahid[1,2], Sheheryar Khan[1], Hong Yan[1]
[1]City University of Hong Kong, 999077 Kowloon, Hong Kong
[2]COMSATS University Islamabad, 45550 Islamabad, Pakistan

## ABSTRACT

Dynamic facial expression recognition has many useful applications in social networks, multimedia content analysis, security systems and others. This challenging process must be done under recurrent problems of image illumination and low resolution which changes at partial occlusions. This paper aims to produce a new facial expression recognition method based on the changes in the facial muscles. The geometric features are used to specify the facial regions i.e., mouth, eyes, and nose. The generic Fourier shape descriptor in conjunction with elliptic Fourier shape descriptor is used as an attribute to represent different emotions under frequency spectrum features. Afterwards a multi-class support vector machine is applied for classification of seven human expression. The statistical analysis showed our approach obtained overall competent recognition using 5-fold cross validation with high accuracy on well-known facial expression dataset.

## 1. INTRODUCTION

Human emotions are expressed by facial muscular movements, leading towards specific human Facial Expression (FE). Recognizing the human facial expressions is still very challenging task for computer vision systems. Automatic facial expression recognition (FER) is an area of analyzing and recognizing emotion characteristics by the means of latest computational devices. FER system application includes: vehicle conversational interface [1], medical and health-care [2] and depressive symptom detection [3].

Human facial expressions have high degree of consistency in facial musculature among people across all cultures. Psychologists established six prototypes, universally recognized expressions: happy (HA), sad (SA), disgust (DI), angry (AN), fear (FE), surprise (SU) [4] (Figure 1). Humans are capable of recognizing the emotions based on their natural intelligence. However, automatic facial expression recognition system is still difficult for computers in the uncontrolled environment such as low light conditions and head pose variations. The six basic expressions are based on movement of facial muscle at particular face locations such as around eyes, nose and mouth. Every particular movement of face muscle is characterized for simplicity known as action units (AUs) [4]. In order to develop an efficient system, facial feature extraction and selecting the appropriate classifier are two key steps. Two key techniques are generally used to extract facial features. First, geometric features based on the edges of specific fiducial points from different facial parts and second, appearance based features to represent the muscular texture of the face using change of wrinkles and creases. Study of the FER methods, applied on geometric based features are Active Appearance Model (AAM) [6], elastic bunch graph matching (EBGM) [7], and straight-line distances [8] of fiducial points. On the other hand, appearance based feature extraction techniques are Local Fisher Discriminant Analysis (LFDA) [9], local binary patterns (LBP) [10] and recently Convolutional Neural Networks (CNN) [11]. The problem with previous implementation of geometrical methods is the accurate localization of facial fiducial points. These methods usually require reliable and accurate facial fiducial point tracking and detection. This is a challenging task in real world applications. On the other hand, appearance based methods are suspectable to lighting condition, head pose and subject dimension.

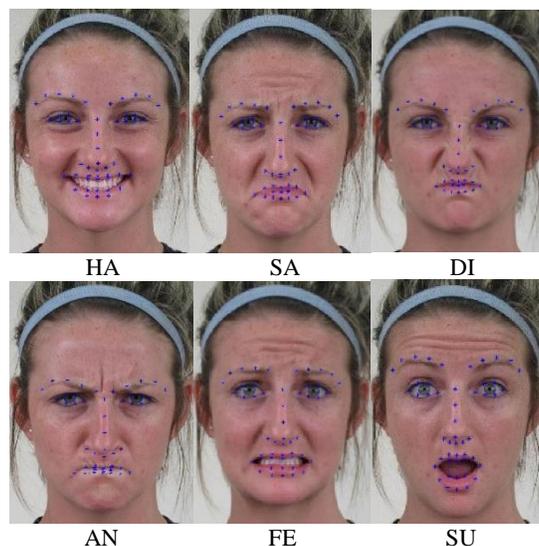

Figure 1. Face landmarks illustration of six basic facial expressions using CFEE dataset [5].

In this research, our proposed method is based on the combination of contour and region shape facial descriptors in Fourier domain. These Fourier descriptors are obtained from geometric features of local and global facial regions i.e., eyes, eye-brows, mouth and nose. Having independent sub-space in the Fourier domain for each facial region overcomes the common FER imaging problems such as registration, illumination condition, dimensionality and redundancy in data. Finally, facial expressions are classified using multi-class support vector machine (SVM) and evaluated using 5-fold cross validation method using the latest dataset for study of facial expressions.

The paper is organized as follows: Section II presents review of shape descriptors. Section III discusses our proposed system including contour and region based shape descriptors and classification. The experiment results are discussed in Section IV, and Section V concludes the paper.

## 2. SHAPE DESCRIPTORS AS FEATURE

To the best of authors' knowledge, the idea for introducing the Fourier shape descriptors to localize features in the geometrical based descriptors feature for facial expression recognition has never been applied before. The process of computing Fourier shape descriptors produces a feature set based on Fourier coefficients. These coefficients form the most discriminative feature vector. Many types of shape features description methods are developed by researchers for recognition task. These descriptors are evaluated on the basis of how accurately shape features allow one to retrieve with the similar shapes from data base (DB) [12]. A robust shape descriptor finds the similar shapes from the DB whether they are finely transformed shapes such as translated, rotated, scaled, flipped, etc.

The shape descriptors are categorized into contour and region based methods. The contour based shape descriptors method uses the boundary information include Fourier Descriptor (FD) [13, 14], wavelet descriptors [15], elliptic Fourier descriptor [16]. On the other hand, the region based shape descriptors uses the pixel information within the shape region. These methods use the whole region into account for shape representation and description includes, Zernike moments [17], grid [18] and generic Fourier descriptor [19]. The effectiveness of such methods is validated through the reconstruction of shape from the computed descriptor and the computational complexity. Extraction of shape feature from human perception is difficult in context of robust shape description. Many shape description methods are proposed in the past. In this regard the more detailed literature review of shape description techniques and analysis could be found in [20].

Nevertheless, all the existing approaches aim for shape feature extraction functionality related perception set under several experiment conditions in machine vision. However, in these approaches, no attention is made to address Fourier shape descriptors selected as features for facial expression classification. In this paper, we propose fused contour and region based Fourier descriptor based feature to effectively utilize the geometrical shape of facial muscles. In addition, choosing the appropriate features for shape recognition system must consider the characteristics of shape that what kind of features are suitable for task. We derive a mechanism for extracting geometric features, especially in FER, where facial expression in contour of mouth and eyes in one direction and the rest of facial characteristics region feature in other direction. The necessity of this mechanism lies in the fact that shape transformed domains are invariant along with the context of translation, rotation, and scaling. Moreover, they are resistant to noise, occultation, and to non-rigid deformation.

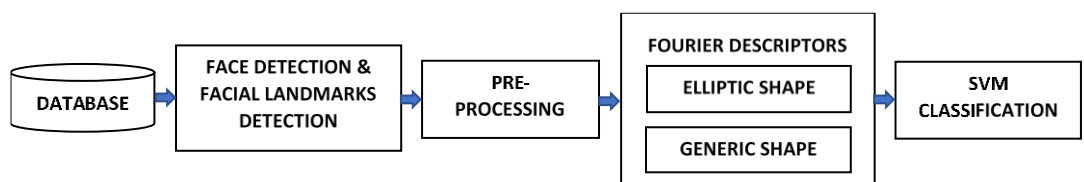

Figure 2. Overview of the proposed method.

## 3. FACIAL EXPRESSION RECOGNITION SYSTEM

In this section, the general framework of our facial expression recognition (FER) system based on Fourier shape descriptors is described. The block diagram of the system is shown in Figure 2. The system consists of three stages: Face and landmarks detection, Facial geometrical feature extraction, and Expression classification. Images are taken from the facial expressions image database, then landmarks and face boundary are extracted form input data. Afterwards the proposed facial geometrical features are extracted and the final stage contains a classifier. A multi-class SVM is applied for recognition task and is also validated using the standard subject independent cross validation schemes. The componential flow of FER system is discussed in the following subsections.

### 3.1 Face and landmarks detection

To align the face features, image normalization is necessary. Although, very good performance for the face and landmark localization has been shown by many researchers [21]. When it comes to the applications that requires an excellent accuracy such as facial behavior and motion. We first detect the face and landmarks using the Chehra face tracker algorithm [22]. This method is based on updating a discriminative model that is trained by cascade of regressors that provide efficient strategy for automatic face and landmarks imaging model for wild conditions. A lot of work has been done in literature which can be accessed in [23].

### 3.2 Facial geometrical feature extraction

#### 3.2.1 Elliptic Fourier shape descriptor

The procedure of elliptic Fourier series approximation involves representation of the spatial coordinate points on the curve in two dimensions. The ellipse estimate is the essential step involved in representing the elliptic type of shape in computer vision [24]. The elliptic shape of mouth and eyes region is estimated using the normalized spatial coordinate values obtained by landmark detector. The normalization, as pre-processing step is applied by dividing difference of the mean of input data with standard deviation of input data. Then geometrical shape of eyes and mouth is estimated by using the ellipse in implicit form and mapped into the general equation as:

$$Ax^2 + By^2 + Cxy + Dx + Ey + F = 0 \qquad (1)$$

where, $A, \ldots, F$ are the constant values, that represents the contoured information. The elliptic Fourier coefficients are then calculated based on an elliptic boundary estimation. The elliptic Fourier series approximation of estimated elliptic contour is then projected on the spatial coordinate as shown in Figure 2. By supposing linear interpolation between adjacent points the of the $n^{th}$ harmonic $a_n, b_n, c_n,$ and $d_n$ can be calculated using the following equations:

$$a_n(t) = \frac{T}{2n^2\pi^2} \sum_{p=1}^{K} \frac{\Delta x_p}{\Delta t_p} \left( \cos\left(\frac{2n\pi t_p}{T}\right) - \cos\left(\frac{2n\pi t_{p-1}}{T}\right) \right)$$

$$b_n(t) = \frac{T}{2n^2\pi^2} \sum_{p=1}^{K} \frac{\Delta x_p}{\Delta t_p} \left( \sin\left(\frac{2n\pi t_p}{T}\right) - \sin\left(\frac{2n\pi t_{p-1}}{T}\right) \right)$$

$$c_n(t) = \frac{T}{2n^2\pi^2} \sum_{p=1}^{K} \frac{\Delta x_p}{\Delta t_p} \left( \cos\left(\frac{2n\pi t_p}{T}\right) - \cos\left(\frac{2n\pi t_{p-1}}{T}\right) \right)$$

$$d_n(t) = \frac{T}{2n^2\pi^2} \sum_{p=1}^{K} \frac{\Delta x_p}{\Delta t_p} \left( \sin\left(\frac{2n\pi t_p}{T}\right) - \sin\left(\frac{2n\pi t_{p-1}}{T}\right) \right) \qquad (2)$$

where, $\{a_n, b_n, c_n, d_n\}$ are determined elliptic Fourier coefficients. The $n$ harmonics required are estimated from average Fourier spectra $fs$ for representing the facial features of elliptic regions as in (3).

$$fs = \frac{\sum_{n=1}^{N}(a_n^2 + b_n^2 + c_n^2 + d_n^2)}{2} \qquad (3)$$

The Fourier coefficient of a harmonic is directly proportional to amplitudes which measures the shape information by harmonic which is computed. $fs$ based on elliptic Fourier descriptor is used as feature of contoured eyes and mouth which is further used as a feature for facial expression classification.

### 3.2.2 Generic Fourier Descriptor

The modified polar Fourier transform is used to derive the Generic Fourier Descriptor (GFD) on the shape of an image. GFD is a region based method for shape description [19]. For the extraction of rest of the salient feature of facial muscles the GFD is used which motivated us by its efficient computation and robustness. Moreover, GFD has also been used to various pattern recognition and computer vision problems [25].

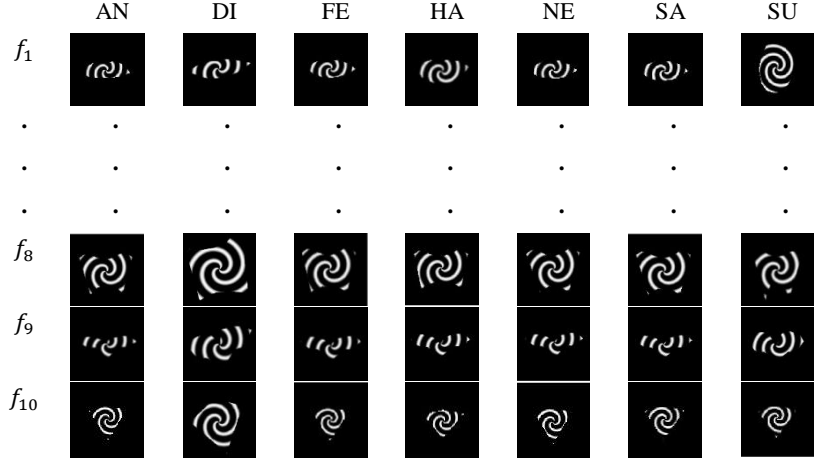

Figure 3. Set of facial geometrical vectors for sub local regions of facial landmarks.

Generally, FD is derived by applying the two dimensional Fourier transform on the shape signature in polar transformed domain. In region based shape descriptor methods, the shape descriptors are computed using the all pixel information within shape region. So, by using the landmarks points as in Figure 1, the binary masks are computed in order to obtain the polar transformation of local regions of face. The region-based descriptors are variant to translation and scaling, so normalization is required to overcome the dimensionality constraint of facial expression and feature extraction process. At first, length in $x, y$ direction of the target shape is obtained from the computed landmarks. Then, image region is constructed based on the landmark values. The cropped image is positioned at the center of a mask which has the same aspect ratio as the input image of shape. The given shape image as:

$$I = f(x,y); 0 < x < m, 0 < y < n$$

$f(x,y)$ is a binary function in shape application. Then shape image $I$ is converted from Cartesian space to $I_p$ polar space.

$$I_p = f(r,\theta); 0 < r < R, 0 < \theta < 2\pi$$

where $R$ is the maximum radius of the shape. The origin of the polar space is set to be the central of the shape, so that the shape is translational invariant. The centroid for $x, y$ coordinates is given by the (4):

$$x_c = \frac{1}{n}\sum_{x=0}^{n-1} x, \ y_c = \frac{1}{m}\sum_{y=0}^{m-1} y \qquad (4)$$

The distance $r$ and angle $\theta$ can be calculated based on the following;

$$r = \sqrt{(x - x_c)^2 + (y - y_c)^2}$$

$$\theta = tan^{-1}\left(\frac{y - y_c}{x - x_c}\right)$$

The $pf(\rho, \phi)$ is the polar Fourier transform (5), $R$ is radial frequency resolution and $T$ is the angular frequency resolution, $m$ is the number of radial frequency maximum and $n$ is the number of angular frequency, the obtained $pf(\rho, \phi)$ is based on the following equation, and the $GFD$ is obtained based on (6).

$$pf(\rho,\phi) = \sum_r \sum_i f(r,\theta_i) \, exp\left[-2\pi j(\frac{r}{R}\rho + \frac{2\pi}{T}\phi)\right] \quad (5)$$

$$GFD = \left\{\frac{pf(0,0)}{2\pi r^2}, \frac{pf(0,1)}{pf(0,0)}, \ldots, \frac{pf(m-1,n-1)}{pf(0,0)},\right\} \quad (6)$$

where, $0 \leq r < R$ and $\theta_i = 2\pi i/T$ $(0 \leq i < T)$; $0 \leq \rho < R, 0 \leq \phi < R$. Figure 3 shows set of ten geometrical regions $\{f_1, \ldots, f_{10}\}$ mask to visualize the selected radial and angular frequencies which is further used to compute the GFD feature vectors.

### 3.3 Expression Classification

Once the features are extracted, the next task is to classify the expression using these features. The SVM classification method is considered to be optimal and effective for classification of the facial expressions. SVM is based on essential risk minimization principle, superior to the traditional empirical risk minimization principle used by neural networks. The working principle of SVM is to transform the input feature vectors to a higher dimensional space by a non-linear transform. It is a learning system that separates the input feature vector in two classes with optimal separating hyperplane [26]. In our work, multi-class SVM one against all strategy is adopted as classifier to recognize the facial expressions. We applied non-linear Gaussian radial basis function (RBF) kernel function for non-linear classification. To reduce the search space of parameter sets, training feature data using the following RBF kernel $k(x,x') = exp[-\gamma\|x - x'\|^2]$. The effectiveness of SVM depends on kernel selection and the optimized parameters. So, each combination of parameter choices is checked using the 5-fold cross validation used to optimize the accuracy of the support vector classification model. Detailed illustration about the multiclass SVMs parameterization for facial expression recognition can be found in [27]. After training the $N$ class support vector machines, a new testing Fourier shape based fused feature vector is classified.

Table 1. Confusion matrix in percentage for best results of SVM classifier (overall accuracy 91.8%).

|    | AN   | NE   | DI   | FE   | HA  | SA   | SU  | *TP*  | *FN*  |
|----|------|------|------|------|-----|------|-----|-------|-------|
| AN | **84.8** | 0    | 12.2 | 3    | 0   | 0    | 0   | *84.8* | *15.2* |
| NE | 3    | **90.9** | 0    | 0    | 0   | 6.1  | 0   | *90.9* | *9.1*  |
| DI | 2.9  | 0    | **91.2** | 2.9  | 0   | 3    | 0   | *91.2* | *8.8*  |
| FE | 3    | 9.1  | 0    | **81.8** | 3   | 3.1  | 0   | *81.8* | *18.2* |
| HA | 0    | 0    | 0    | 0    | **100** | 0    | 0   | *100*  | *0*    |
| SA | 0    | 2.9  | 5.9  | 8.8  | 0   | **79.5** | 2.9 | *79.5* | *20.5* |
| SU | 0    | 0    | 0    | 0    | 0   | 0    | **100** | *100*  | *0*    |

## 4. EXPERIMENTS

This section includes the comparison and analysis of FER system based on the proposed shape descriptor as feature strategy. In our experiments, we focused on Compound Facial Expressions of Emotions (CFEE) [5]. The CFEE database consist of 230 human subjects including both human genders with average age of 23. It includes most of ethnicities and races around the world. In total consist of 5060 images annotated with prototypical AUs for each basic and 22 compound emotion categories. The image size is $3000 \times 4000$ with RGB variation. We have selected 1610 images from CFEE database which corresponds to the seven classes (i.e., neutral plus six basic emotions). The confusion matrix shows the accuracy of the classifier individually for the six basic expressions and neutral (Table I). The performance of happy and surprise expression is very accurate in comparison to the rest of expressions because they are easy to recognize. Although angry, fear and sad are suspectable to misclassification for particular subjects because of resemblance. The overall accuracy achieved by the multiclass SVM classifier is 92%. In Table I, the TP (True positive) and FN (False negative) rate shows the overall confusion rate of expressions. If the problem is reduced to six basic expressions as considering the neutral is no emotion the accuracy results improved to 94% in context of six class facial expression classification.

## 5. CONCLUSION

We have developed a shape description method for facial expressions based on the Fourier shape description features. Our experiments demonstrate the selected features have a high classification power in terms of FER rate. A subset of fused shape descriptors as feature sets are used to classify the seven basic facial expressions, using the radial basis function kernel multi-class support vector machine (SVM) classifier. Where the simulation results revealed that the shape descriptor approach extracts the low dimensional and discriminant feature space for geometrical facial muscles representation. The proposed method is suitable for implementing FER systems to wild data due to sub local facial description. The selected subset shape features not only lead to better recognition accuracy but also reduces the computational time complexity significantly in context of dimensionality and illumination condition.

## ACKNOWLEDGEMENT

This work is supported by the Hong Kong Research Grants Council (Project C1007-15G) and City University of Hong Kong (Project 9610034).